# REGULARIZATION AND FALSE ALARMS QUANTIFICATION: TWO SIDES OF THE EXPLAINABILITY COIN

**Nima Safaei**  NIMA.SAFAEI@SCOTIABANK.COM
*Data Science & Analytics (DSA) Laboratory*
*Bank of Nova Scotia*
*40 King St. West*
*Toronto, M5H 1H1, Canada*

**Pooria Assadi**  POORIA.ASSADI@CSUS.EDU
*College of Business Administration,*
*California State University*
*6000 J Street,*
*Sacramento, CA 95819, USA*

## Abstract

Regularization is a well-established technique in machine learning (ML) to achieve an optimal bias-variance trade-off which in turn reduces model complexity and enhances explainability. To this end, some hyper-parameters must be tuned, enabling the ML model to accurately fit the unseen data as well as the seen data. In this article, the authors argue that the regularization of hyper-parameters and quantification of costs and risks of false alarms are in reality two sides of the same coin, explainability. Incorrect or non-existent estimation of either quantities undermines the measurability of the economic value of using ML, to the extent that might make it practically useless.

## 1 Introduction

Machine Learning (ML) technologies are transforming the practice of management. However, for every article that suggests to executives and managers that they can win using ML and artificial intelligence, there is one that warns them of the challenges ahead. Hence, business executives and management teams in various industries are increasingly facing the question of why and when they should put ML-based technologies to work. In this article, we offer a new, practical business case for when managers might find ML-based applications useless versus useful. In doing so, we highlight the importance of the quantification of false alarms in any meaningful cost-benefit analysis of the usefulness of ML-based products and services. We do so because the quantification of false alarms at the organization level is directly tied to tuning the regularization hyper-parameters, which in turn is tied with reducing ML complexity and improving ML explainability. By doing so, we show that regularization and quantification of costs and risks of false alarms are in reality two sides of the ML explainability coin, such that any incorrect or non-existent estimation of false alarms undermines the measurability of the economic value of using ML to the extent that it might make ML practically useless.

### 1.1 Why is ML used?

Perhaps the most common answer to the question of why ML is used is to reduce the human intervention and speed up the decision making process. As such, the power of an ML model depends on how well it can fit the unseen data (in addition to the seen data) with a high level of accuracy. The learning theory suggests that improving the power of ML models requires achieving an optimal bias-variance trade-off, which in turn demands reducing overfitting and underfitting at the same time. A strategy that is often used to achieve an optimal bias-variance trade-off is regularization.

### 1.2 Regularization and the Mystery of Hyper-Parameter Optimization

Regularization is an enforcement mechanism in the training phase of ML models that limits or shrinks the learning coefficients to help achieve an optimal bias-variance trade-off and reduce model *complexity*. Reducing model complexity in turn leads to improving model *explainability* because research has shown that there exists a trade-off between model complexity and explainability (Har, 2019). To reduce model complexity or equivalently improve explainability, regularization imposes a limit on the magnitude of model parameters (Lever et al., 2016; Poggio et al., 2020).

Regularization is applicable to all parametric ML models (e.g., regression) and some non-parametric models which can be formulated as constrained/combinatorics optimization problems such as deep networks (Amos & Kolter, 2017; Poggio et al., 2020), support vector machines (Xu et al., 2016), LASSO problem (Bertsimas & King, 2016), and Tree Ensembles (Mišić, 2020).

More formally, given a generic regression or supervised classification problem, the regularization efforts seek to reduce the complexity of the hypothesized learning model $Y = f(X, Y; W)$ where $W$ represents the vector of learning coefficients; given that, $\Gamma = \{(X_1, Y_1), (X_2, Y_2), \ldots, (X_t, Y_t), \ldots (X_T, Y_T)\}$ represents the historical observations with $X$ as feature set and $Y$ as dependent variable or label. Regularization is applied using penalty function $g(W)$ and penalty coefficient $\lambda$, so-called *regularization hyper-parameter*.

The objective of learning is to minimize $\ell(f(X; W), Y) + \lambda g(W)$ over all possible subsets of observations (i.e., training sets) where $\ell$ is a generic loss function (e.g., mean squared error, cross-entropy, hinge, or logistic). As Figure 1 shows, $g(W)$ is usually presented in form of L1-norm or LASSO ($\sum_i |w_i|$), L2-norm or Ridge or Tikhonov ($\sum_i w_i^2$), or a weighted combination of both known as L1/2 or Elastic Net form (Hui & Trevor, 2005; Zhang et al., 2020). If $W$ is desired to be sparse with few nonzeros (i.e., a desired goal in the high dimensional learning frameworks), $g(W)$ may be represented as L0-norm or $\sum_i sgn(|w_i|)$ (Hazimeh & Mazumder, 2020). The latter is the case for the subset selection problem in the field of combinatorics optimization.

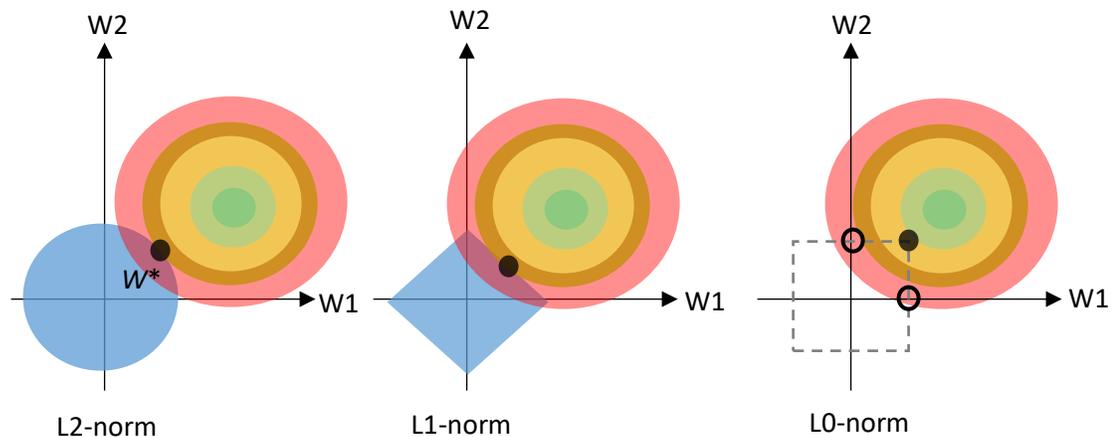

**Figure 1.** Different forms of regularization

For illustration, let the learning model $f$ represent a regression model. As it relates to regularization, the learning model that uses L1-norm regularization is referred to as Lasso regression, and the learning model that uses L2-norm regularization is referred to as Ridge regression (Khalaf & Shukur, 2006). The primary difference between these techniques is that Lasso shrinks the less important features' coefficients to zero, therefore, removing some features altogether. This works well for feature selection scenarios where there are a large number of features. In addition, in this case, the learning model that weighs learning coefficients differentially is refereed to as Heteroskedastic Ridge regression. Lastly, the learning model that combines the penalties of Lasso and Ridge regressions (i.e., to get the best of the two settings) is referred to as Elastic Net regression (Hui and Trevor, 2005).

Minimizing $\ell(f(X;W),Y) + \lambda g(W)$ involves determining regularization hyper-parameter $\lambda$ which may be a single entity or a vector. For instance, in Gaussian kernel SVM, $\lambda = (C, \delta)$ is a hyper-parameter vector where coefficient $C$ is regularization penalty to control the margin and $\delta$ is bandwidth of Gaussian kernel (Feng & Liao, 2017). In a similar fashion, in deep learning models, the learning rate along with the number of layers and neurons per layer are hyper-parameters that need to be optimally determined.

To optimize the regularization hyper-parameter, whether a single entity or a vector, *stochastic-guided interior search* strategy is often utilized, including the grid search (Hutter, 2009), random search (Bergstra & Bengio, 2012), Gaussian process (Bergstra, et al. 2011), stochastic gradient decent (Poggio et al., 2020), evolutionary algorithms, and Bayesian inference (Mardani, 2020). However, there are serious concerns associated with using these techniques.

First, the regularization penalty function itself results in a more complicated and mostly a non-convex loss function, which makes the optimization process very costly and challenging. Second, the abovementioned techniques do not guarantee the optimal solution because the search space is often uneven with multiple local optima. Third, due to their stochastic nature, these techniques require running multiple trials in a single training job, which in turn makes them very time-consuming and complicated to tune. Fourth, the answer to the important and practical question of "how well do these techniques perform?" is, in reality, *not* supported by strong theoretical foundations and is instead problem-dependent. It is then not surprising that the majority of related research on this topic answered this question empirically.

Because of these concerns, the hyper-parameter optimization remains a mysterious aspect of ML models. In this paper, we argue that the primary reason for this mystery is that we do not yet have a clear business/economic *interpretation* for regularization hyper-parameter $\lambda$.

## 2    Interpreting Regularization Hyper-parameter Using Lagrange Multiplier

One potential way of interpreting the regularization hyper-parameter $\lambda$ is to reconsider the learning objective function as a Lagrangian function. To do so, the learning problem can be rewritten as a constrained optimization problem $\min_{(X,Y)} \ell(f(X;W),Y)$ subject to $g(W) \leq c$ where $c$ is a *complexity control* quantity that serves as an upper-bound for the penalty function. In this setting, then, the associated Lagrangian function is $L(W,\lambda) = \ell(f(X;W),Y) + \lambda(g(W) - c)$ where the non-negative factor $\lambda$ is known as the Lagrange multiplier. The critical point of interest in this setting $(W^*, \lambda^*)$ is the optimal solution to $\nabla L(W,\lambda) = 0$. In particular, as Figure 2 illustrates, $\lambda^*$ reveals how much more improvement on the loss function one could make by modifying the upper bound $c$.

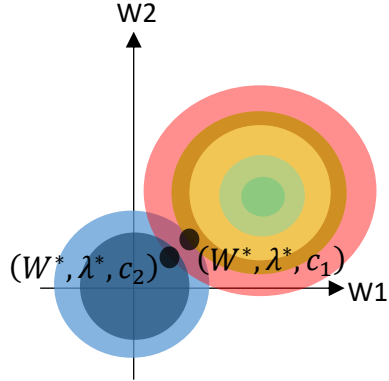

**Figure 2.** Lagrangian interpretation of regularization

However, a serious shortfall of a Lagrangian interpretation is that another quantity, *complexity control c*, is required to interpret the regularization hyper-parameter $\lambda$. Yet, $c$ itself is another hyper-parameter to be determined and it significantly depends on the form of the penalty function.

## 3   Reinterpreting the Regularization Hyper-Parameter

Using a simple and intuitive mathematical manipulation, in this paper, we offer a new interpretation for the regularization hyper-parameter $\lambda$ to address the conundrum of hyper-parameter optimization, one that does not need determining additional quantities. Since $\lambda$ is a positive constant, without loss of generality, we rewrite the learning objective function as $min\,\gamma \ell(f(X;W),Y) + g(W)$, where $\gamma = \frac{1}{\lambda}$ is the *unit cost or risk of model's error or false alarm*, that is, the cost or risk of misclassification, a wrong prediction, or an incorrect decision.

Note that $\gamma$ may represent a vector of false components. For example, in a binary classification setting, $\gamma = (\gamma_{FP}, \gamma_{FN})$ represents the unit cost or risk of both False Positive (FP) and False Negative (FN) components. However, in a multi-class classification setting, $\gamma$ can be represented in matrix form $[\gamma_{ij}]$—similar to the confusion matrix—such that the entity $\gamma_{ij}$ is the unit cost or risk when an observation known to be in group $i$ is predicted to be in group $j$ where $\gamma_{ii} = 0\,\forall i$.

The unit cost or risk of false alarms manifest themselves differently across various applications. For example, in the real-world applications where anomaly detection is critical, false negatives tend to be costly and false positives tend to be risky. In recommendation systems, false positives may translate into *opportunity cost*. In settings such as aviation systems, airplane components, fire alarms, transaction fraud discovery in financial systems, and autopilot systems of self-drive vehicles (Brown, 2019), failure to detect anomalies have proven to be particularly risky.

To illustrate our proposed reinterpretation, let $Y = f(X,Y;W)$ represent a recommendation engine that is designed to determine a set of targeted customers to whom a new product should be recommended. In this context, then, the unit cost of a false positive $\gamma_{FP}$ is the operational and dissatisfaction costs associated with a customer declination even though the trained model classified that customer as a potential target, and the unit cost of a false negative $\gamma_{FN}$ is the opportunity cost of missing a potential customer. Indeed, $\gamma_{FP}$ and $\gamma_{FN}$ reveal the unit cost of incorrect detection because of the gap between the training and implementation phases, a gap that is often caused by the ignorance on prior knowledge and/or potential missing features. This example, and its underlying interpretation, can be extended to other settings beyond recommendation engines.

## 4 When Is ML Practically Useless?

With our proposed interpretation, we argue that ML is not particularly useful for applications where either the unit cost or risk of error cannot be practically quantified (i.e., $\gamma \to \infty$ or $\lambda \to 0$) or the unit cost or risk of error is negligible (i.e., $\gamma \to 0$ or $\lambda \to \infty$). In the former, the risk or cost is practically intolerable, and in the latter the risk or cost is irrelevant (i.e., so small that makes no difference).

As an example of a case where $\gamma \to \infty$, consider an unsupervised ML model responsible for detecting the potential failures in aircraft engines (Cottrel, et al. 2009). Even though such model's accurate performance is highly desirable, it comes with a high risk of false alarms. In particular, any false positive could be very costly due the consequences associated with aircraft downtime (Safaei & Jardine, 2018). In addition, the risk of false negatives could be extremely high due to the catastrophic failures or fatality crashes resulting in loss of life (Safaei, 2019). In such a scenario, decision makers cannot solely rely on the performance of ML models, and instead must consult with expert humans ensure that model outcomes are consistent with safety and reliability standards and regulations. In fact, in scenarios like this, there is often no real opportunities for reducing human intervention and speeding up the decision making process using ML.

As an example of a case where $\gamma \to 0$, consider movie suggestion platforms (e.g., Netflix or Hulu). In such platforms, where the unit cost or risk of false alarms is practically small, it is often not useful to target specific customers to maximize profits. In applications like this, the least complex (most conservative) approach will prove to be more useful, that is, to recommend the new product (e.g., new movie) to all customers.

## 5 When Is ML Useful?

We posit that ML is particularly useful for applications where the unit cost or risk of error or false alarm is practically quantified and nonnegligible (i.e., $0 < \gamma \ll \infty$). In the scenario involving recommendation engines, suppose that $\gamma = (\$A, \$B)$ where $\$A$ is the unit opportunity cost of each false negative per customer and $\$B$ is the unit cost of each false positive per customer (e.g., personalized advertisement, marketing campaign, skilled labor, information technology infrastructure, ML deployment). In addition, suppose that the probability of a false negative (probability of ignoring a potential customer) is $\alpha\%$ and the probability of a false positive (probability of targeting a wrong customer) is $\beta\%$. Further suppose that the out-of-sample size (number of targeted customers) is $N$ and the product price (i.e., revenue earned per acquired customer) is $\$C$ where $\$A = \$C - \$B$.

Absent ML, the optimal strategy would be to recommend the new product to all customers at the cost of $N \times B$. This strategy will produce a total revenue earning of $N \times C \times \vartheta$ where $\vartheta$ is the probability of a customer's acceptance of the recommendation having adopted a non-ML approach. The total net profit that this strategy generates, then, is $N(C\vartheta - B)$.

Given these assumptions, an ML model would be useful only if the total net profit earned as a result of using ML minus the cost of false positives is greater than the total net profit earned absent ML. That is, $N_1(C(1 - \beta) - B) - N_0 \alpha B > N(C\vartheta - B)$. Note that $1 - \beta$ is the probability that a customer is truly predicted as targeted customer. This means that an ML model would be useful only if $N_1 C(1 - \beta) > NC\vartheta - N_0 B(1 - \alpha)$ or equivalently $N_1 C(1 - \beta) + N_0 B(1 - \alpha) > NC\vartheta$ where $N_0$ is the predicted number of false-class (i.e., non-targeted) customers and $N_1$ is the predicted number of true-class (i.e., targeted) customers in the out-of-sample dataset where $N = N_0 + N_1$.

The latter inequality shows that for ML to be useful, the total revenue earned due to ML, $N_1 C(1 - \beta)$, plus the total cost saved by ML, $N_0 B(1 - \alpha)$, must be greater than the total revenue earned absent ML, $NC\vartheta$. In other words, if the expected revenue per targeted customer is equal to expected saving per non-targeted customer, i.e., $C(1 - \beta) = B(1 - \alpha)$, then ML would be useful if the probability of true positive of ML (i.e., *precision*) is greater than the precision of using any non-ML approach (i.e., probability of customer's acceptance of a recommendation), i.e., $1 - \beta > \vartheta$. Our proposed reinterpretation of $\gamma$ lends support to such practical analysis of usefulness of ML.

For numerical illustration, Figure 3 shows the contour plot of the surface associated with the ML net profit by varying $0 < \beta < 1$ and $0 < C < 100\$$ where $N_0 = 70$, $N_1 = 30$, $\vartheta = 0.4$, $B = 5\$$ and $\alpha = 0.05$.

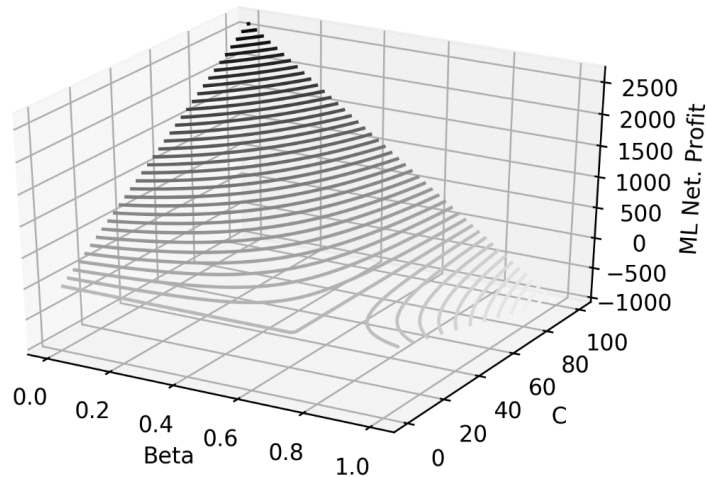

**Figure 3.** Contour plot of ML profitability

## 6   Conclusion

The majority of real-world ML applications lack cost-benefit analysis of their usefulness. One key reason for this deficiency is lack of understanding and quantification of false alarms. At its core, we argue that quantification of false alarms at the organization level is directly tied to tuning the regularization hyper-parameter, which in turn is correlated with reducing ML model complexity and improving ML model explainability.